\documentclass[runningheads]{llncs}

\usepackage{eccv}

\usepackage{eccvabbrv}
\definecolor{mypink}{RGB}{239,43,159}
\usepackage{graphicx}
\usepackage{booktabs}

\usepackage[accsupp]{axessibility}  % Improves PDF readability for those with disabilities.

\usepackage{hyperref}

\usepackage{orcidlink}
\usepackage{xcolor} % 引入xcolor包

\newlength\savewidth

\begin{document}

\title{Towards Natural Language-Guided Drones: 
\texorpdfstring{GeoText-1652 Benchmark with Spatial \\ Relation Matching}
{GeoText-1652 Benchmark with Spatial Relation Matching}}

\titlerunning{Towards Natural Language-Guided Drones}

\author{Meng Chu\inst{1} \orcidlink{0000-0002-8125-1308}\and
Zhedong Zheng\inst{2}\thanks{Corresponding author.}\orcidlink{0000-0002-2434-9050}  \and
Wei Ji\inst{1} \orcidlink{0000-0002-8106-9768}\and \\
Tingyu Wang\inst{3}  \orcidlink{0000-0002-4169-1595}
\and Tat-Seng Chua\inst{1} \orcidlink{0000-0001-6097-7807}
}

\institute{School of Computing, National University of Singapore, Singapore \and
FST and ICI, University of Macau, China
\and
School of Communication Engineering, Hangzhou Dianzi University, China\\
\email{  e0998106@u.nus.edu, zhedongzheng@um.edu.mo,\\
\{jiwei, dcscts\}@nus.edu.sg,  tingyu.wang@hdu.edu.cn } \\
\color{mypink} \textbf{ \url{https://multimodalgeo.github.io/GeoText/}}
}

\authorrunning{Chu et al.}
\maketitle
\begin{abstract}
Navigating drones through natural language commands remains challenging due to the dearth of accessible multi-modal datasets and the stringent precision requirements for aligning visual and textual data.
To address this pressing need, we introduce GeoText-1652, a new natural language-guided geolocalization benchmark. This dataset is systematically constructed through an interactive human-computer process leveraging Large Language Model (LLM) driven annotation techniques in conjunction with pre-trained vision models. 
GeoText-1652 extends the established University-1652 image dataset with spatial-aware text annotations, thereby establishing one-to-one correspondences between image, text, and bounding box elements.
We further introduce a new optimization objective to leverage fine-grained spatial associations, called blending spatial matching, for region-level spatial relation matching. 
Extensive experiments reveal that our approach maintains a competitive recall rate comparing other prevailing cross-modality methods. This underscores the promising potential of our approach in elevating drone control and navigation through the seamless integration of natural language commands in real-world scenarios.
\end{abstract}

\section{Introduction}
\label{sec:intro}

Drone navigation using natural language offers potential to a range of applications such as disaster management \cite{mehbodniya2022improving,9155522}, live search and rescue \cite{meguro2006disaster,brunsting2016geotexttagger}, and remote sensing
\cite{huang2019flight,hu2023geographic,chandarana2017fly,blukis2019learning}.
Given one single input image, drone navigation is to search the other relevant images of the same place from a large-scale gallery \cite{zheng2020university,zhang2023cross, shi2019spatial,10129939,rodrigues2021these,yu2019building}, which 
 is usually regarded as a sub-task of image retrieval. 
Current datasets typically provide pairs of images, focusing on matching images from disparate platforms like drones and satellites \cite{radford2021learning,li2021align,jia2021scaling,zhang2020context,zhu2021detection,wang2022multiple}. However, the query image is not always available, while natural language description is a more intrinsic input modality from the user. There remain two challenges to natural language-guided drone navigation:
(1) There is no large public language-guided dataset. Providing such a detailed description of the image is usually challenging with high human resource costs and reliable annotation quality. 
(2) It remains difficult to align language and visual representation due to the fine-grained nature of the drone-view scene images.

%-------------------------------------------------------------------------

\begin{figure}[t]
	\centering
	\includegraphics[width=0.9\linewidth]{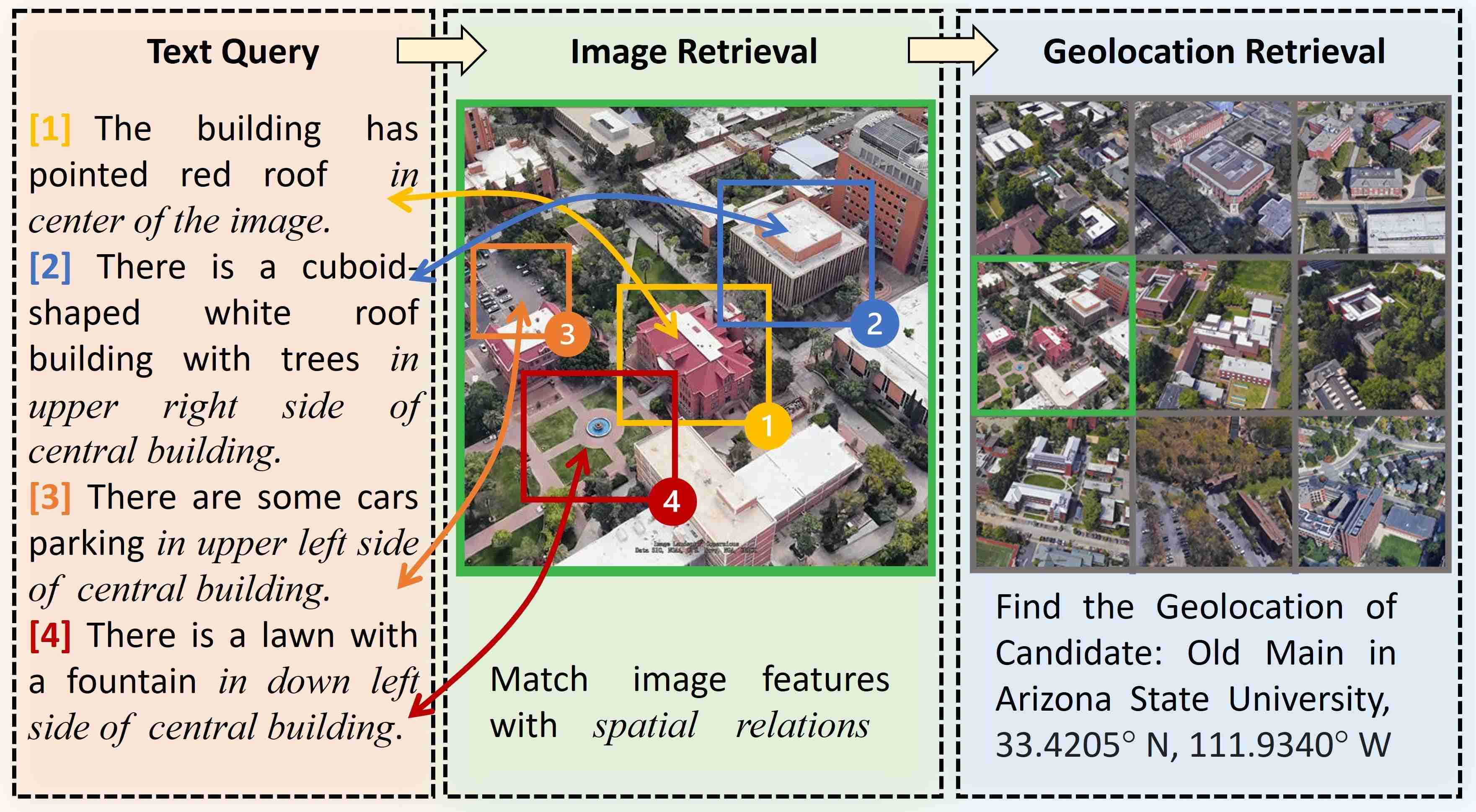}
       \vspace{-.1in}
	\caption{\textbf{An example of the proposed benchmark, GeoText-1652.} Here we show a text-guided drone geolocalization process. Left: Every image contains several region-level query sentences.  Middle: Given the user description, we match the text and region of interest with the spatial relation. Right: With the dense spatial relation matching, we could easily retrieve the place of interest against other similar false-positives, and navigate the drone. \textbf{It is worth noting that multiple similar-appearance buildings usually exist in the neighbour regions, so we also indicate the relative position, \eg, left, right, upper, and down, in the text query.}  }
	\vspace{-1.5em}
	\label{fig:intro}
\end{figure}

For the first limitation, we propose a multi-view, multi-source vision-language dataset GeoText-1652, drawing on the existing multi-source images dataset, University-1652~\cite{zheng2020university} (see Fig.~\ref{fig:intro}). We have established links between locational data and corresponding textual descriptions through a semi-automatic procedure, annotation including  276,045 text-bbox pairs and 316,335 descriptions.
Our benchmark facilitates two new tasks: drone navigation via text and drone-view target localization.  
As the name implies, drone navigation via text focuses on the strategic guidance of a drone to the location it has previously visited that most closely aligns with a provided textual description. This involves a
fine-grained text-to-image retrieval process, highlighting the integration between linguistic and spatial data. 
On the other hand, drone-view target localization 
focuses on identifying the textual description that best matches a drone-captured image to accurately localize a target, which is an image-to-text retrieval task. 
In our experimental setup, we approach these tasks as challenges in cross-modal retrieval focusing on bridging different types of data representations. We compare the generic feature trained on extremely large datasets with the viewpoint-invariant feature learned on our proposed dataset.  
We observe that the proposed GeoText-1652 dataset aids in learning the viewpoint-invariant feature, which refines drone control via language, making it more precise and intuitive. To address the second challenge, we introduce an approach for spatial relation matching that leverages the GeoText-1652 dataset. Our methodology encompasses two losses, grounding loss and spatial loss, which help the model in understanding the spatial relationships between objects within images. Through this approach, we enhance the capability of the model to decipher spatial correlations for more precise text-to-image retrieval. The main contributions are as follows:

%that explores the region-level relationship between drone images and text. %In particular, we focus on establishing a connection between position and text annotation using human-computer interaction annotation.
\begin{itemize}  
\item In pursuit of facilitating natural language-guided drone geolocalization, we 
introduce a new image-text-bbox benchmark, called GeoText-1652, which builds upon the existing multi-platform University-1652 image dataset. Our key contribution lies in establishing precise associations between spatial positions and their corresponding text annotations through an innovative human-computer interaction-based annotation process.

\item As a minor contribution, we propose a new spatial-aware approach that leverages fine-grained spatial associations to perform region-level spatial relation matching. Different from the independent bounding box regression, our approach further introduces relative position within drone images and textual descriptions of surrounding positions to achieve precise localization.

\item Our proposed spatial-aware model has achieved 31.2\% recall@10 accuracy using text query, surpassing established models, such as ALBEF \cite{li2021align}, and X-VLM \cite{zeng2022multi}. 
Moreover, our model shows promising generalization capabilities when applied to unseen real-world scenarios, % it has not been previously exposed to, 
highlighting its potential for effective use in diverse and unseen environments.

\end{itemize}

\section{Related Works}
\textbf{Cross-view Geolocalization.}
Cross-view geolocalization addresses the challenge of associating images captured from different viewpoints with their corresponding geographical locations \cite{zheng2020university,wang2023generalized,berton2022deep,jin2017learned,trivigno2023divide}. One key underpinning this task is to extract a discriminative visual representation against viewpoints. For instance, Wang~\etal \cite{wang2021each} develop a partitioning strategy that enriches the feature set by considering multiple parts of the image and Lin~\etal~\cite{lin2022joint} introduce a new attention module to discover representative key points and focus on the salient region. Dai~\etal~\cite{dai2021transformer}  introduce a transformer-based structure with a content alignment strategy. 
Similarly, Yang~\etal~\cite{yang2021cross} utilize the properties of self-attention and exploit the positional encoding of ground and aerial images.  
Rodrigues~\etal \cite{rodrigues2022global} introduce a dual path network to fuse the local region with the global feature for partial aerial-view image matching. 
Another line of works~\cite{shi2022beyond,zhang2023cross,dhakal2023sat2cap,Zhu_2022_CVPR,hu2022beyond,chen2023cross} further integrate enhanced features across different model designs and leverage extra knowledge to improve geolocalization. Shi~\etal \cite{shi2022beyond} 
fuse the pose estimation and geometry projection into the feature matching, while Hu~\etal \cite{hu2022beyond} emphasize the accuracy of orientation in street-view images. Chen~\etal \cite{chen2023cross} introduce a cross-drone mapping mechanism in the transformer. GeoDTR~\cite{zhang2023cross} employs two data augmentation techniques to capture both low-level details and spatial configurations. TransGeo~\cite{Zhu_2022_CVPR} combines transformer flexibility and attention-guided non-uniform cropping to enhance image resolution in key areas. Dhakal~\etal \cite{dhakal2023sat2cap} design one contrastive learning framework, which could predict textual embedding for ground-level scenery. Different from existing methods, our work focuses on two new natural language-guided drone tasks, which provide a straight-forward way to control the drone.

\noindent\textbf{Multi-modality Alignment.}
In this work, we focus on natural language-guided navigation, which can be viewed as a sub-task of
text-to-image retrieval \cite{radford2021learning, li2022blip,dou2022empirical}. Early works usually focus on structure design, such as dual-path network~\cite{zheng2020dual}. Wang~\etal~\cite{wang2019camp} employ an adaptive gating scheme to handle negative pairs and irrelevant information, calculating the matching score based on the fused features, while Li~\etal \cite{li2019visual} apply graph convolutional networks for semantic reasoning within image regions. Then, Chen~\etal~\cite{chen2020uniter} propose word region alignment in the pertaining of multi-modal model with large-scale datasets. Li~\etal \cite{li2020oscar} apply object tags detected in images as anchor points to ease the learning of alignments, while Yang~\etal~\cite{yang2023towards} study the attribute-related keywords. Clip model~\cite{radford2021learning} proposes a contrastive learning method between image-text pairs. Jia~\etal \cite{jia2021scaling} design a simple dual-encoder architecture to align visual and language representations. Li~\etal~\cite{li2021align} refine image text matching loss with a self-training method which learns from pseudo-targets.
Zeng~\etal \cite{zeng2022multi} further align multi-region visual concepts and associated texts. Blip model~\cite{li2022blip} leverages noisy web data through a caption bootstrapping process. 
Different from these existing works, we introduce a spatial-aware approach, which explicitly considers fine-grained spatial text-region matching. %For instance, when similar elements appear within a single image, the model may incorrectly align them if it fails to discern the nuances in their spatial relations.

\noindent\textbf{Data Synthesis via Large Models.}
Deep learning-based automatic annotation has thrived in recent years\cite{pasquini2021automated, vaucher2020automated}. Drawing from the success of AI Generated Content (AIGC), numerous studies have harnessed the capabilities of Large Models (LM)\cite{zhao2023survey} for the creation of training or supplementary datasets.
The LM has already showed the ability to do annotation for different modality data, including the text\cite{kuzman2023chatgpt, Gilardi_2023, zhang2023llmaaa}, image\cite{li2023stablellava,ikezogwo2024quilt}, video\cite{shvetsova2023howtocaption, maaz2023video}, and music\cite{doh2023lpmusiccaps, liu2023music}.  Wang~\etal \cite{wang2023generative}, utilizing LM to tailor personalized content for recommendation systems. Concurrently, Hämäläinen~\etal \cite{hamalainen2023evaluating} delve into GPT-3's capacity to craft credible user research narratives for human-computer interaction. This trend extends into the domain of enhancing data precision and utility, with endeavours such as Yu~\etal \cite{yu2023generate} exploration of LMs in generating open-domain QA content and Meng~\etal \cite{meng2023tuning} produce synthetic data for few-shot learning boosting classification task performance. The collective progress in this field, from the generative capabilities demonstrated by Borisov~\etal \cite{borisov2022language} in tabular data synthesis to Fang~\etal \cite{fang2023domain} %'s creation of
synthetic molecules, reflects that champions not just the generation of data but its thoughtful curation and refinement to meet the nuanced demands of various tasks. Chen~\etal \cite{chen2024allava} harness the GPT4V-synthesized data to build a lite vision-language model. This paradigm shift, further propelled by methodological enhancements such as those proposed by Yu~\etal \cite{yu2024large} for curating less biased and more diverse training datasets.
However, our approach diverges significantly from these predecessors by offering more detailed annotations and tailoring our methodology specifically for spatial matching tasks. This nuanced focus not only enhances the granularity of the data provided but also optimizes the dataset for more precise and effective application in spatial analysis, setting a new precedent in the utilization of LMs for dataset synthesis.

\noindent\textbf{Vision and Language Navigation.}
Vision-and-Language Navigation (VLN) requires an agent to follow natural language instructions to navigate in a specific environment\cite{anderson2018vision}. Recent approaches tackle VLN using cross-modal attention\cite{majumdar2020improving,qi2021object}, data augmentation~\cite{thomason2020vision, zhu2020vision}, and incorporating object-level information and structured spatial representations \cite{georgakis2022cross, hong2021vln, qi2021object}. Memory architectures \cite{hong2021vln, majumdar2020improving}, auxiliary reasoning tasks\cite{zhu2020vision}, pre-training on image-text pairs \cite{hao2020towards}, and integrating referring expressions \cite{majumdar2020improving, qi2021object} have shown promise in improving navigation. In this work, we focus on the language-guided drone navigation task, which remains under-explored.

\begin{figure*}[t]
    \centering
    \includegraphics[width=0.98\linewidth]{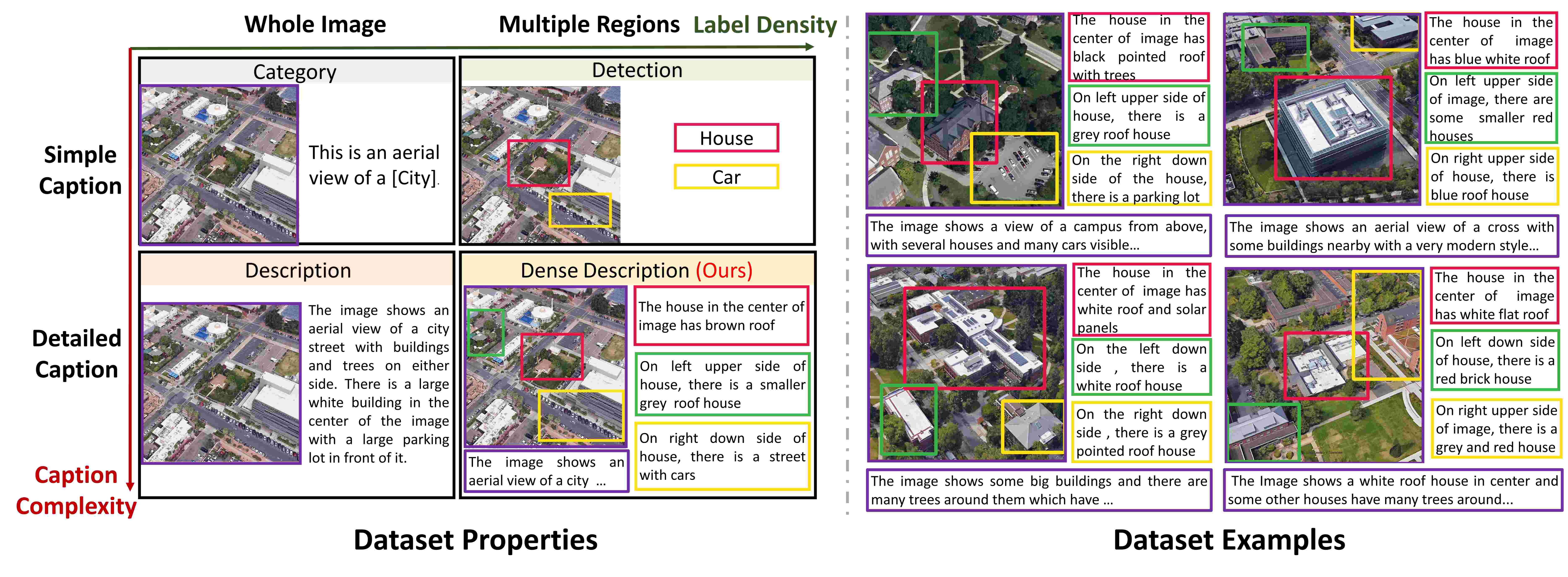}
    \vspace{-.1in}
    \caption{
\textbf{The properties of the proposed dataset GeoText-1652.} Different from the traditional category annotation, our dataset not only
    includes image-level detailed descriptions
    but region-level short descriptions (left). Samples of the dataset show that the description could align well with the image and its regions (right).}
    \label{fig:data}
\end{figure*}

\setlength{\tabcolsep}{6pt}
\begin{table}[t]
%\fontsize{8pt}{11pt}\selectfont
\scriptsize
\begin{center}
\begin{tabular}{c|c|c|c|c|c}
\toprule
Split & \#Imgs & \#Global Descriptions & \#Bbox-Texts  & \#Classes & \#Univ.\\
\hline
Training$_{drone}$ & 37,854 & 113,562& 113,367 & 701 &  \multirow{3}{0.06\linewidth}{\centering{33}} \\
Training$_{satellite}$ & 701 & 2,103 & 1,709 & 701 &  \\
Training$_{ground}$ & 11,663 & 34,989 & 14,761 & 701 &  \\
\hline

Test$_{drone}$& 51,355 & 154,065 & 140,179  & 951 & \multirow{3}{0.06\linewidth}{\centering{39}}\\
Test$_{satellite}$ & 951 & 2,853 & 2,006 & 951 & \\
Test$_{ground}$ & 2,921 & 8,763 & 4,023 & 793 & \\
\bottomrule
\end{tabular}
\end{center}
\vspace{-.1in}
\caption{\textbf{Statistics of GeoText-1652.} Training and test sets all include the image, global description, bbox-text pair and building numbers. 
%The test set only uses the image and global description during the test process. 
We note that there is no overlap between the 33 universities of the training set and the 39 universities of the test sets. Three platforms are considered, \ie, drone, satellite, and ground cameras.
}
\label{table:dataset_table}
\vspace{-.1in}
\end{table}

\section{GeoText-1652 Dataset} \label{sec:dataset}
\subsection{Dataset Description}
The proposed GeoText-1652 dataset extends the image-based University-1652 dataset~\cite{zheng2020university}, containing 1,652 buildings in 72 universities from three platforms, \ie, satellite, drone and ground cameras. %All images are annotated 
We add fine-grained annotations for every image with 3 global descriptions and 2.62 bounding boxes on average since we removed some low-quality bounding boxes. Specifically, each global description, encompassing both image-level and region-level details, contains 70.23 words on average. 
As shown in  Fig.~\ref{fig:data}, the proposed dataset, compared to the original dataset, contains fine-grained descriptions with region-level annotations, which is the key to the natural language-guided task. The region-level descriptions, extracted specifically for bounding box matches, contain 21.6 words on average. More detailed statistics are shown in Table~\ref{table:dataset_table}. 
Similar practices in other computer vision fields, \eg, those by Zhu~\etal~ \cite{zhu2024multimodal} and COCO-Captions\cite{chen2015microsoft}, also affirm that enriching single modal datasets with visual or textual data could enhance model training for fine-grained vision-language tasks.

\subsection{Dataset Annotation Framework}
\begin{figure}[t]
    \centering
    \includegraphics[width=0.98\linewidth]{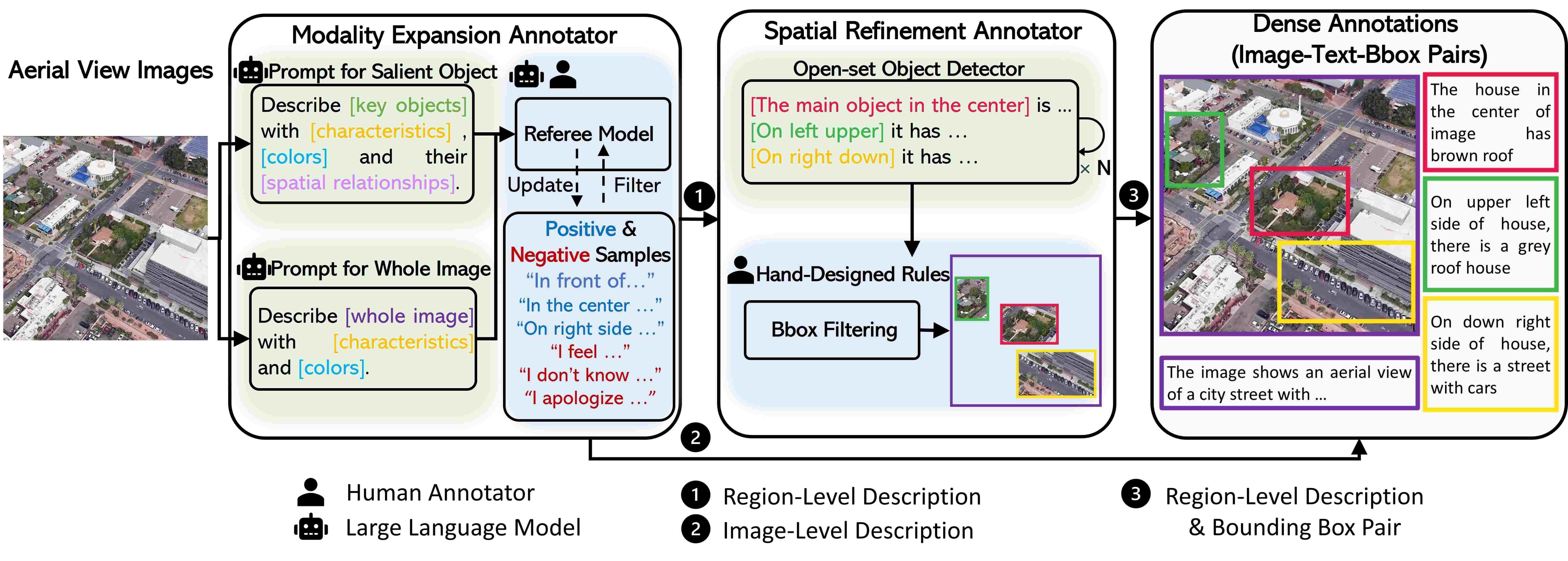}
    \vspace{-.1in}
    \caption{\textbf{The proposed human-computer interaction annotation strategy.} The strategy includes two main processes: modality expansion annotator and spatial refinement annotator. The modality expansion annotator is to annotate the image-level and the region-level descriptions. 
    The spatial refinement annotator could utilize the region-level description to conduct the visual grounding. Finally, after human-computer filtering processes, we build the proposed dataset with Image-Text-Bbox Pairs.}
    \vspace{-.1in}
    \label{fig:dataset_annotation}
\end{figure}

As shown in Fig.~\ref{fig:dataset_annotation}, we briefly
illustrate the overall workflow of our dataset construction for the natural language-guided geolocalization. 
We extend the conventional drone-view dataset University-1652 with dense annotations.
To generate image-text pairs, we adopt a new human-computer interaction annotation strategy, which could largely save time and costs. % adopted by some exsited human-LLM collaborative datasets \cite{sun2023medmmhl,zheng2023lmsyschat1m,kamalloo2023hagrid}.
Considering LLMs still have problems in reasoning, including diverse biases, hallucinatory responses, and inconsistencies, even for advanced models such as GPT-4V\cite{chen2024mllmasajudge}, we argue that human validation is of importance during the process~\cite{pangakis2023automated}.
In particular, our annotation process has two principal phases, \ie, the modality expansion phase and the spatial refinement phase.

%curated through human-computer interaction. It includes 
\noindent\textbf{Modality Expansion Phase.} For the modality expansion phase, we apply two kinds of prompts for each image. One prompt focuses on salient objects, and the other prompt encompasses the description of the entire image. Given the input and prompts, we ask the visual language model (visual-LLM~\cite{zhu2024minigpt}) to generate answers. Considering the inherent limitations of language models, such as hallucination phenomena and ambiguous statements, not all outputs meet the standards. 
To address this limitation, we introduce a referee model to autonomously adjudicate whether the outputs from visual-LLM meet good quality. 
The raw answers (1) undergo positive sample element detection to ensure the inclusion of the desired keywords and (2) then enter a negative sample pool to exclude subjective statements. 
The keyword within the referee model is set by the human-computer interaction. In particular, given several raw answers, \ie, 1,000 cases, we adopt another large language model~\cite{openai2023gpt4} as a teacher to classify the negative and positive samples. 
The referee model keyword list is updated with 
 terms in negative samples that typically indicate common errors, like `img src', `[image]', and various apologies or URLs. 
In contrast, the positive word list is for spatial relationship indicators
to ensure the inclusion of positional context within the annotations.
The human annotator only needs to check the key word list. If the presence of negative terms or missing positive words triggers the referee model, the visual-LLM will re-generate the caption until meeting the standard.
This process enables us to obtain three image-level descriptions and nine region-level description proposals for each input image.

\begin{figure*}[t]\centering
\vspace{-1em}
\centering\subfloat[
\label{fig:scale}
]{
 \begin{minipage}{0.57\textwidth}
        %\fontsize{0.1pt}{5pt}\selectfont
        \begin{center}
\resizebox{0.95\linewidth}{!}{
\begin{tabular}{l|c|c|c|c}
\toprule
Property & CVUSA\cite{workman2015localize} & CVACT\cite{Liu_2019_CVPR} & VIGOR\cite{zhu2021vigor} & GeoText-1652 \\ \hline
Annotation & GPS Tag & GPS Tag & GPS Tag & Sentence \\
\# Bbox-Texts & N/A & N/A & N/A & 276,045 \\
Platform & G,S & G,S & G,S & G,S,D \\
Modality & Image & Image & Image & Image, Text \\ 
\bottomrule
\end{tabular}}
        \end{center}
        \vspace{-0.1in}
        \label{table:dataset_comparision}
    \end{minipage}
    }
\hspace{-1em}
\subfloat[]{
 \begin{minipage}[b]{0.4\linewidth}
\includegraphics[width=0.98\linewidth]{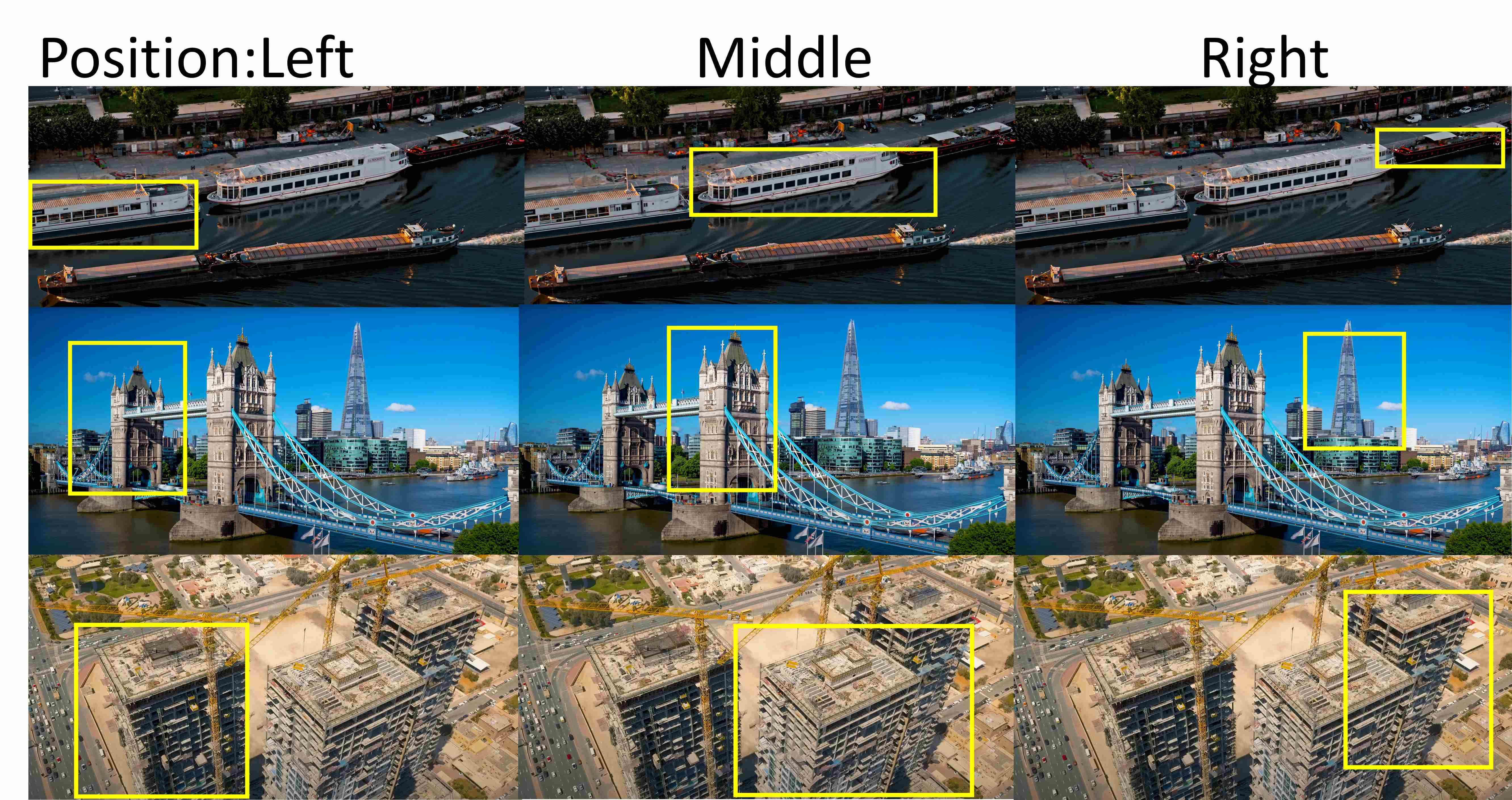}
\label{fig:features_demo}
\end{minipage}
}
\vspace{-.5em}
\caption{\textbf{(a)}%\textbf{Dataset Comparison:}
Comparison between the proposed GeoText-1652 dataset and other existing geolocalization datasets. The labels G, S, and D represent ground-view, satellite-view, and drone-view images, respectively. \textbf{(b)}\textbf{Why is Relative Position Necessary?}
Here we show some typical challenging cases.
In these rows, similar objects (boats, towers, skyscrapers) are difficult to distinguish based on their characteristics alone. However, their spatial relationships (left, middle, right) can effectively aid in distinguishing them.
Color-coded boxes highlight the main object from left to right.
}
\label{ablation}
\end{figure*}

\noindent\textbf{Spatial Refinement Phase.} In the spatial refinement phase, we build the relationship with the bounding box. Specifically, given the region-level description, we utilize an off-the-shelf text-based visual grounding model\cite{liu2023grounding} to identify corresponding bounding boxes (bboxes). Since all region-level descriptions contain spatial phrases such as `right' or `left', we set a spatial rule to filter out the bounding boxes in mismatched locations. We also refine the description by adding vertical spatial terms like `upper left' and `down right'. 
Considering the domain gap between pre-trained grounding models and our aerial-view data,  we empirically fine-tune the inference hyper-parameters, \eg, IoU threshold, in grounding models via feedback. We conduct 5-round evaluation.  In each round, we randomly extract 20\% of the annotations for human evaluation, assessing both the accuracy of the bounding box and the relevance of the associated text. These evaluations are graded on a matching scale. 
Over five iterations of refinement, the annotations rated as excellent exceed 90\% according to manual feedback.   
We only preserve the high-quality 2.62 corresponding bounding boxes and region-level description for every image on average. Finally, through the stages of modality expansion and spatial refinement annotation, we achieve dense annotations, encompassing both image-level descriptions and region-level descriptions with bbox pairings. This iterative and multi-faceted approach ensures a high-quality dataset for fine-grained geolocalization using natural language.

\noindent\textbf{Discussion. The contribution to the community.}
The key difference from existing datasets \cite{Liu_2019_CVPR,zhu2021vigor,workman2015localize} lies in the fine-grained region-level descriptions, facilitating more intuitive natural language-guided tasks (see Fig.~\ref{table:dataset_comparision}). This level of detail is crucial for tasks requiring precise localization and contextual understanding. For instance, only describing the visual patterns of the main building can be challenging due to language limitations (see  Fig.~\ref{fig:features_demo}). 
In such cases, distinguishing the target by describing the surrounding buildings can be an effective strategy. 
The spatial context adds clarity and distinction. Enhancing a multimodal model with relative spatial reasoning is crucial for interpreting fine-grained visual contexts.
Moreover, our dataset annotation framework, incorporating a human-computer interaction strategy and a referee model, ensures both efficiency and high-quality annotations. 
Researchers can leverage GeoText-1652 to explore new approaches, improve model generalization, and push the boundaries of visual geolocalization and natural language understanding integration.

\section{Method}

\begin{figure*}[t!]
    \centering
    \vspace{-.1in}
    \includegraphics[width=0.98\linewidth]{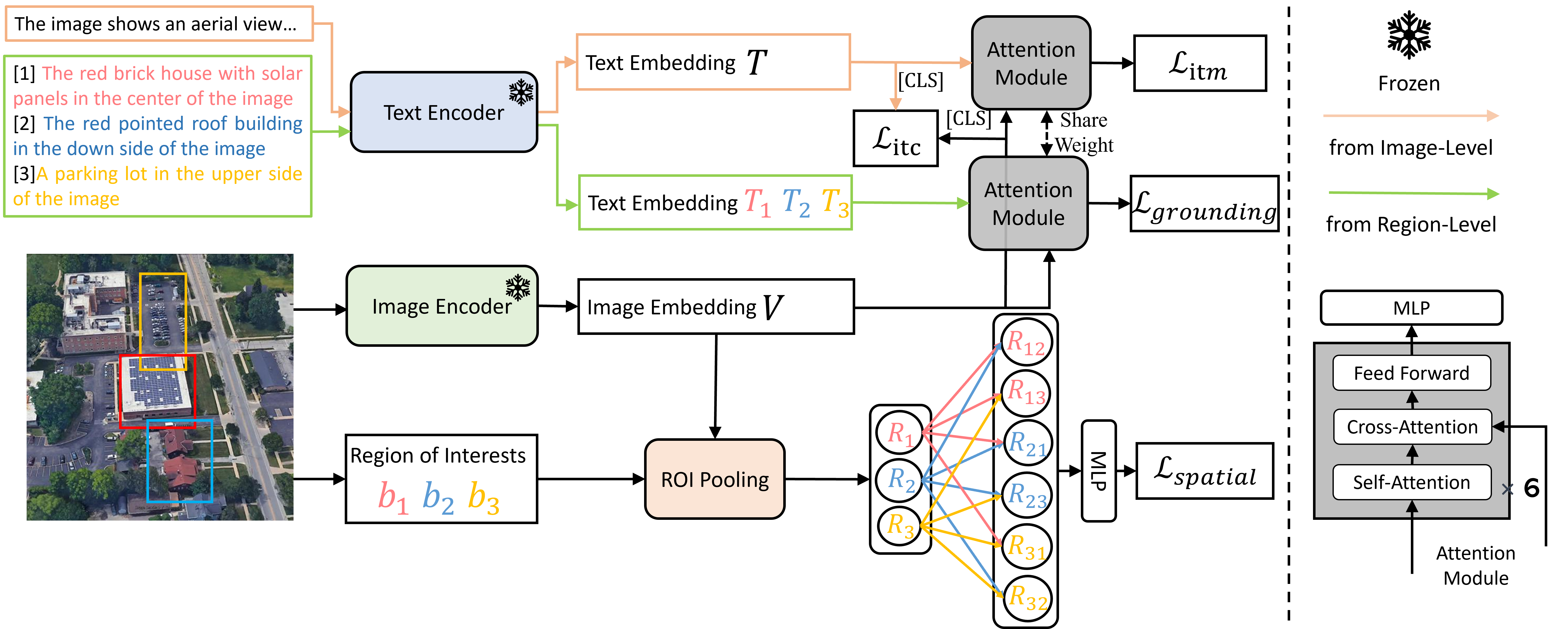}
    \normalsize
    \vspace{-.15in}
    \caption{\textbf{The proposed multi-modal framework.} The framework processes an aerial image by identifying regions of interest (ROIs) and matching them with corresponding text descriptions. It contains an image encoder that extracts visual embeddings and intermediate feature maps. 
    We could obtain region-level visual features via ROI Pooling, and concatenate to calculate the spatial relation followed by multi-layer perceptron (MLP). 
    On the other hand, text inputs, including the image-level and region-level descriptions, are encoded separately with the text encoder. 
    Two attention modules integrate the image and text features, and they share the same weights. 
    The framework applies several loss functions, including Grounding and Spatial Loss for blending spatial matching, and ITM and ITC Loss for image-text matching. }
    \label{fig:model_architexture}
\end{figure*}

We introduce a cross-modal geolocalization framework to conduct fine-grained spatial analyses %, merging visual and textual data
(see Fig.~\ref{fig:model_architexture}). It mainly consists of an image encoder, a text encoder, and a cross-modal encoder. %All encoder structures are based on transformer~\cite{vaswani2017attention}.
We revisit the image-text semantic matching in Sec.~\ref{sec:image-text matching}, followed by the new blending spatial matching in Sec.~\ref{sec:blending spatial matching}.

\subsection{Image-text Semantic Matching}
\label{sec:image-text matching}

\noindent\textbf{Image-Text Contrastive.} Given an image-text pair, we first extract the image visual feature $V$ and image-level text feature $T$, respectively.  
Cosine similarity can be calculated as: $s(V,T) = \frac{V^\top T}{||V||_2 ||T||_2}$. 
According to contrastive learning, we treat the other samples within the mini-batch as negative examples. 
Then, we could calculate the in-batch vision-to-text and text-to-vision similarity as: 
\begin{align}
\boldsymbol{p}_\mathrm{v2t} = \frac{\exp (s(V,T) / \tau)}{\sum_{i=1}^N \exp (s(V,T^i)/ \tau)}, \quad 
%\label{eq:pi2t}
\boldsymbol{p}_\mathrm{t2v} = \frac{\exp (s(V,T)/ \tau)}{\sum_{i=1}^N \exp (s(V^i,T)/ \tau)},
\end{align}
where $\tau$ is a learnable temperature parameter. The superscript of $V^i$ and $T^i$ denotes the $i$-th sample within the batch. The contrastive learning is defined as: %the cross-entropy loss % between $\boldsymbol{p}$ and $\boldsymbol{y}$:
\begin{equation}
\mathcal{L}_\mathrm{itc} = -\frac{1}{2} \mathbb{E} \left[  \log(\boldsymbol{p}_\mathrm{t2v})+ \log(\boldsymbol{p}_\mathrm{v2t}) \right],
\end{equation}
where we encourage that the identity image-text pair has the larger similarity. 

\noindent\textbf{Image-Text Matching.} We further demand the model to determine whether a pair of visual concepts and text is matched. For each visual concept in a mini-batch, we sample an in-batch hard negative text feature according to the highest similarity in Eq.~\ref{eqn:itm}.
Similarly, we also sample one hard negative visual feature for each text. We apply the output embedding of the cross-modal encoder to predict the matching probability $\boldsymbol{p}_\mathrm{match}$, and the binary classification loss is:
\begin{equation}
\label{eqn:itm}
\mathcal{L}_\mathrm{itm} = -\mathbb{E}\big[\boldsymbol{y_m}\log(\boldsymbol{p}_\textrm{match})+(1-\boldsymbol{y_m})\log( 1-\boldsymbol{p}_\textrm{match})\big],
\end{equation}
where $\boldsymbol{y_m}$ is a binary label indicating whether the input is a positive pair or a hard negative pair. % the ground-truth label.

\subsection{Blending Spatial Matching}
\label{sec:blending spatial matching}
In text-guided bounding-box prediction, known as the grounding process, the model uses natural language descriptions to identify and spatially locate objects within an image. This involves an interaction between the text and image feature maps %, where the model uses the semantic information from the text
to guide a region proposal network. Therefore, our proposed blending spatial matching includes two optimization objectives: the grounding prediction and the spatial relation matching. 

\noindent\textbf{Grounding Prediction.}
Given the image representation and the region-level text representation, the model is to predict the bounding box $\boldsymbol{b}_{j}$ according to the corresponding textual concept $T_j$. The bounding box is formulated as $\boldsymbol{b}_{j}=(c_x, c_y, w, h)$.  Here the subscript $j$ denotes the $j$-th short bounding box of the corresponding image. $c_x, c_y$ are the center point coordinate of the bounding box, and $h, w$ are the height and width, respectively.
In particular, we adopt the cross-attention model with six transformer blocks followed by multi-layer perceptron (MLP) (as shown in Fig.~\ref{fig:model_architexture} (right)). We also apply the Sigmoid to normalize the prediction $\hat{\boldsymbol{b}}_{j}$ within the valid region $[0, 1]$. 
The grounding prediction loss includes the $\ell_1$ regression loss and the Intersection over Union (IoU) loss~\cite{rezatofighi2019generalized} to compare the overlap areas. Therefore, the grounding loss can be formulated as: 
\begin{equation}
\mathcal{L}_\mathrm{grounding} = \mathbb{E} [\mathcal{L}_\mathrm{iou}(\boldsymbol{b}_{j}, \hat{\boldsymbol{b}}_{j}) + ||\boldsymbol{b}_{j}- \hat{\boldsymbol{b}}_{j}||_1 ].
\end{equation}

\noindent\textbf{Spatial Relation Matching.} 
Considering the grounding loss focus on a single region, we propose a relative localization matching. For instance, 
given the visual feature of three bounding boxes, we %can compute 
intend to predict the spatial relationship between them.
Given three regions of interests \(b_1\), \(b_2\), \(b_3\), we extract the visual feature based on the global feature $V$ via the ROI Pooling module as region features \(R_1\), \(R_2\), \(R_3\). 
As the spatial relation is a relative concept, we concatenate the region features as composed feature \(R_{ij}\) ($i \neq j$).  Then we adopt the Multi-Layer Perceptron (MLP) to predict the 9-class spatial relationship $\boldsymbol{p_r}^{ij}$. 
The spatial loss is defined as the cross-entropy loss between $\boldsymbol{y_r}^{ij}$ and $\hat{\boldsymbol{p_r}}^{ij}$:
\begin{equation}
\label{eqn:spatial}
\mathcal{L}_\mathrm{spatial} = \mathbb{E}[-\boldsymbol{y_r}^{ij} \log(\hat{\boldsymbol{p_r}}^{ij})],
\end{equation}
where the ground-truth class $\boldsymbol{y_r}^{ij}$ is derived by the center distance for the two bboxes  \((c_x, c_y, w, h)\) and \((c_x', c_y', w', h')\). Horizontal distance is defined as \(\Delta x = c_x' - c_x\) and vertical distance is \(\Delta y = c_y' - c_y\). 
If \(|\Delta x| < \frac{w}{2}\), we define it as `middle'; 
If  \(\Delta x > \frac{w}{2} \), we define it as `left'; 
If  \(\Delta x < -\frac{w}{2} \), we define it as `right'.
Similarly, we also could classify the ground-truth vertical relationship as  3 categories, \ie, top, middle, and bottom. Therefore, we could compose the vertical and horizontal relation as 9 location categories in total. 

\noindent\textbf{Discussion. Why do we need spatial relation matching?} Relative position estimation has been explored in other fields, such as self-supervised learning~\cite{doersch2015unsupervised}. 
In this work, spatial matching serves as a crucial complement to bounding box prediction in our approach, providing a nuanced perspective on the relationships between different regions of interest (ROIs). While bounding box prediction $\mathcal{L}_\mathrm{grounding}$ focuses on individual regions, our proposed relative localization matching $\mathcal{L}_\mathrm{spatial}$ introduces a relative spatial dimension to the scene understanding. In particular, the proposed spatial relation matching via 9 orientation classification motivates the model towards a more fine-grained understanding of different regions within the image. 

\noindent\textbf{Optimization Objectives.} Finally, the total loss \(\mathcal{L}_{total}\) is defined as:
\begin{equation}
\mathcal{L}_{total} =\mathcal{L}_\mathrm{itc} + \mathcal{L}_\mathrm{itm} + \lambda(\mathcal{L}_\mathrm{grounding} + \mathcal{L}_\mathrm{spatial}),
\end{equation}
where \(\lambda\) is the blending spatial matching weight, and we empirically set \(\lambda = 0.1\).

\section{Experiment}

\noindent\textbf{Implementation Details.} We adopt XVLM~\cite{zeng2022multi} pretrained on 16M images as our backbone model. Our text encoder is BERT~\cite{devlin2018bert} and our image encoder is Swin~\cite{liu2021swin}.  We deploy AdamW~\cite{loshchilov2017decoupled} optimizer with a weight decay of 0.01. The learning rate is set to $3e^{-5}$. All images are resized to 384 $\times$ 384 pixels during the training process, and the image patch size is set to 32. We perform simple data augmentation, such as brightness adjustment and identity operation. We do not use random rotation or horizontal flipping as it would lose the spatial information. In the context of global description serving as the text query, we remove stop words to keep the query concise during evaluation.

\subsection{Geolocalization Performance} \label{sec:localization}

The GeoText-1652 dataset contributes to the advancement of cross-modality retrieval and the proposed method
outperforms the performance of other models, particularly when fine-tuned with this dataset. 
We could observe two primary points from Table~\ref{table:sota}:

\noindent\textbf{Effectiveness of the Proposed Dataset.} The GeoText-1652 dataset, provides a substantial ground for evaluating the image-text retrieval capabilities of various models. 
The results show that fine-tuning our dataset leads to considerable improvements in performance, as seen with ALBEF$_{\textit{finetuned}}$ and XVLM$_{\textit{finetuned}}$, among others. The result also suggests that the dataset contains rich and varied annotations that are beneficial for training models to understand and match images with text descriptions accurately. Furthermore, the significant gap between pre-trained models and their fine-tuned counterparts underscores that it remains challenging for the ``large'' vision model on the aerial-view dataset, reflecting the necessity of the proposed dataset.

\noindent\textbf{Superiority of the Proposed Method.} Our method shows a clear superiority over other methods, particularly in the Recall@10 metric for both image and text retrieval tasks. Compared with the baseline XVLM$_{\textit{finetuned}}$, the proposed method moves more positive candidates forward in the ranking list, with +0.4\% Recall@1, +0.9\% Recall@5 and +1.6\% Recall@10 improvements in the text-to-image retrieval. Similarly, we could observe the increase in the image-to-text retrieval setting, with +1.3\% Recall@1, +1.4\% Recall@5 and +1.8\% Recall@10.
Such improvement is non-trivial in practical applications, where multiple correct answers are desirable. 
With comparable model parameters, the high Recall@10 performance also implies that the model is capable of understanding the visual-textual relationship effectively. 
The proposed approach learns diverse features from the  GeoText-1652 dataset, handling the detailed descriptions and region-level annotations efficiently.

\begin{table*}[!t]
    \vspace{-.1in}
    \scriptsize
    \centering	
    \setlength\tabcolsep{1pt}
    \begin{tabular}	{ c|c|c| ccc|ccc}
    \toprule	
    \multirow{2}{*}{Method} &\multirow{2}{*}{\# Params}& \multirow{2}{*}{\# Pretrained Images} %& \multicolumn{6}{c}{GeoText-1652 } \\\cmidrule(rl){4-9} 
     & \multicolumn{3}{c|}{Text Query}& \multicolumn{3}{c}{Image Query} \\
	 %\midrule
	& & & R@1 &R@5&R@10& R@1 &R@5&R@10\\
 	 \midrule
        UNITER~\cite{chen2020uniter} & 300M & 4M & 0.9 &  2.7 & 4.2& 2.5 & 7.4 & 11.8\\
 	METER-Swin~\cite{dou2022empirical} & 380M & 4M & 1.3 &  3.9 & 5.8& 2.7 & 8.0 & 12.2\\

  	ALBEF~\cite{li2021align} & 210M & 4M & 1.8 &  4.8 & 7.1& 2.9 & 8.1 & 12.4\\
    	ALBEF~\cite{li2021align} & 210M & 14M & 1.1 & 3.5&  5.3& 3.0 & 9.1 & 14.2\\

        XVLM~\cite{zeng2022multi} & 216M & 4M & 4.3 & 9.1& 13.2 & 4.9& 14.2 & 21.1\\
 	XVLM~\cite{zeng2022multi} & 216M & 16M & 4.5 & 9.9& 13.4 & 5.0& 14.4 & 21.4\\
  	 \midrule
        UNITER$_{\textit{finetuned}}$ & 300M & 4M & 10.6 &  20.4 & 26.1& 21.4 & 43.4 & 59.5\\
 	METER-Swin$_{\textit{finetuned}}$ & 380M & 4M & 11.3 & 21.5 & 27.3& 22.7& 46.3 & 60.7\\

  	ALBEF$_{\textit{finetuned}}$ & 210M & 4M & 12.3 & 22.8 & 28.6& 22.9& 49.5 & 62.3\\
    	ALBEF$_{\textit{finetuned}}$ & 210M & 14M & 12.5 & 22.8 & 28.5 & 23.2 & 49.7 & 62.4\\

    XVLM$_{\textit{finetuned}}$ & 216M & 4M & 13.1 & 23.5 & 29.2 & 23.6& 50.0 & 63.2\\
    \hline
 	XVLM$_{\textit{finetuned}}$ & 216M & 16M & 13.2 & 23.7 & 29.6 & 25.0& 52.3 & 65.1\\
  	Ours & 217M & 16M & \textbf{13.6} & \textbf{24.6}& \textbf{31.2} & \textbf{26.3} &  \textbf{53.7} & \textbf{66.9} \\
 		\bottomrule
	\end{tabular}	
      \caption{\textbf{Image-text bi-direction retrieval results on GeoText-1652.} Text Query: Drone Navigation (Text-to-Image Search). Image Query: Drone-view Geolocalization (Image-to-Text Search).  We adopt Recall@K as the evaluation metric.}\vspace{-.1in}
      \label{table:sota}
\end{table*}

\subsection{Ablation Studies and Further Discussion} \label{sec:ablation}

\noindent\textbf{Effect of Loss Objectives.} 
We gradually add the loss terms to train the model, and the retrieval performance is shown in Table~\ref{table:loss_ablation}. 
The baseline model, stripped of both the spatial and grounding loss, exhibits a significant impairment, as mirrored in the Recall@1 accuracy for Text Query and Image Query. With the grounding loss only, the overall performance of the model is better compared to the baseline model, \ie, +0.3\% Recall@1 accuracy in Text Query and +0.9\% Recall@1 accuracy in Image Query. With the spatial loss only, the model shows a consistent enhancement in performance compared to the baseline model, \ie, +0.2\% Recall@1 accuracy in Text Query and +0.3\% Recall@1 accuracy in Image Query. With our method, the evaluation result shows a notable increase, \ie, +0.4\% Recall@1 in Text Query and +1.3\% Recall@1 in Image Query. Therefore, the full model has arrived at the best performance with the two losses together, \ie, 13.6\% Recall@1 in Text Query and 26.3\% Recall@1 in Image Query. We observe that grounding loss is the main factor in enhancing retrieval performance. The combination of both losses performs better than only using one of them.

\noindent\textbf{Different Training Sets.} We study the effect of the dataset split in Table~\ref{table:comparison}. The ``Satellite + Drone + Ground'' training set shows better performance than only using ``Drone'' or ``Satellite + Ground'', \ie, +0.7\% Recall@1 in Text Query and +0.6\% Recall@1 in Image Query compared to ``Drone'' training set, and +3.5\% Recall@1 in Text Query and +7.6\% Recall@1 in Image Query compared to ``Satellite + Ground'' training set. These results indicate that the training set includes a more diverse range of data (for instance, a combination of satellite, drone, and ground data), facilitating the model training.

\begin{table}[t]
\caption{Ablation studies on: \textbf{(a)} Spatial and bbox losses. \textbf{(b)} Different training sets. \textbf{(c)} The hyper-parameter $\lambda$ selection. \textbf{(d)} Rotation angles.}
\vspace{-.15in}
\begin{subtable}{.55\linewidth}
\caption{} \label{table:different_uncertainty}\vspace{-.05in}
\centering
\resizebox{0.9\linewidth}{!}{
\begin{tabular}{c|ccc|ccc}
\toprule
\multirow{2}{*}{Method} & \multicolumn{3}{c|}{Text Query} & \multicolumn{3}{c}{Image Query} \\
& R@1 & R@5 & R@10 & R@1& R@5 & R@10 \\
\midrule
Baseline~\cite{zeng2022multi} &13.2 & 23.7& 29.6 & 25.0& 52.3 & 65.1 \\
\emph{w} grounding loss &13.5 & 24.4& 30.9 & 25.9& 53.4 & 66.3 \\
\emph{w} spatial loss&13.4 & 24.0& 30.1 & 25.3& 52.8 & 65.6 \\
Ours &\textbf{13.6}&\textbf{24.6}&\textbf{31.2}&\textbf{26.3}&\textbf{53.7} & \textbf{66.9}\\
\bottomrule
\end{tabular}
\label{table:loss_ablation}
}
\hfill
\centering
    \caption{} \vspace{-.05in}
    \resizebox{0.9\linewidth}{!}{
\begin{tabular}{l|c|ccc|ccc}
\toprule
\multirow{2}{*}{Training Set} & \multirow{2}{*}{\#imgs}&\multicolumn{3}{c|}{Text Query} & \multicolumn{3}{c}{Image Query} \\
& &R@1 & R@5 & R@10 & R@1& R@5 & R@10 \\
\midrule
Drone & 37,854&12.9 & 23.4 & 29.1 & 25.7 & 51.5 & 64.3 \\
Satellite + Ground & 12,364&10.1 & 19.3 & 24.4 & 18.7 & 39.6 & 51.2  \\
Satellite + Drone + Ground & 50,218&\textbf{13.6} & \textbf{24.6} & \textbf{31.2} & \textbf{26.3} & \textbf{53.7} & \textbf{66.9} \\
\bottomrule
\end{tabular}
    }
    \label{table:comparison}
\end{subtable}
\begin{subtable}{.4\linewidth}
\centering
\caption{}\vspace{-.05in}
\small
\label{tab:gamma}\resizebox{0.98\linewidth}{!}{
\begin{tabular}{c|ccc|ccc}
\toprule
\multirow{2}{*}{$\lambda$} & \multicolumn{3}{c|}{Text Query} & \multicolumn{3}{c}{Image Query} \\
& R@1 & R@5 & R@10 & R@1& R@5 & R@10 \\
\midrule
1.00 &10.5 & 21.6& 27.6 & 21.8& 47.5 & 60.8 \\
0.50 &11.2 & 22.3& 29.4 & 23.2& 51.4 & 63.5 \\
0.10&\textbf{13.6}&\textbf{24.6}&\textbf{31.2}&\textbf{26.3}&\textbf{53.7} & \textbf{66.9}\\
0.05&12.3&24.1&30.6&24.6&52.9&65.2\\
\bottomrule
\end{tabular}
\label{table:lambda}
}
\centering
\caption{}\vspace{-.05in}
\small
\label{tab:gam}\resizebox{0.9\linewidth}{!}{
\begin{tabular}{c|ccc|ccc}
\toprule
\multirow{2}{*}{Rotation Degree} & \multicolumn{3}{c|}{Ours} & \multicolumn{3}{c}{Baseline} \\
& R@1 & R@5 & R@10 & R@1& R@5 & R@10 \\
\midrule
0 &\textbf{13.6}&\textbf{24.6}&\textbf{31.2}& 13.2& 23.7 & 29.6\\
\hline
15 & 13.4 & 24.3 &30.9 & 13.0 & 23.6 & 29.4  \\
90 & 13.1& 23.7 &29.6& 12.9 & 23.4 &  29.1 \\
180&13.3&23.9&30.2& 13.1 & 23.6 & 29.5\\
270&13.2&23.8&29.8& 12.9 & 23.5 & 29.2\\
\bottomrule
\end{tabular}
\label{table:rotation}
}
\end{subtable}
\vspace{-.15in}
\end{table}

\noindent\textbf{Hyperparameter Study.} 
$\lambda$ is the weight to balance the spatial matching losses and the cross-modality matching losses. As shown in Table \ref{table:lambda}, we could observe that when $\lambda=0.1$, the learned mode achieves the best recall rate. 

\noindent\textbf{Rotation Angle Study.} 
We rotate test images at 15°, 90°, 180°, and 270°. As shown in Table \ref{table:rotation}, we observe that the proposed method is robust to small rotation perturbation. As expected, it performs worse against a larger rotation degree, considering that we provide a ``wrong'' relative position in the text query. In contrast, the baseline method achieves a similar performance against rotation. Since the main objects are still correct in the text query, the performance drop is within an acceptable range, and our method still surpasses the baseline. 

\begin{figure*}[t]
    \centering
    \includegraphics[width=0.98\linewidth]{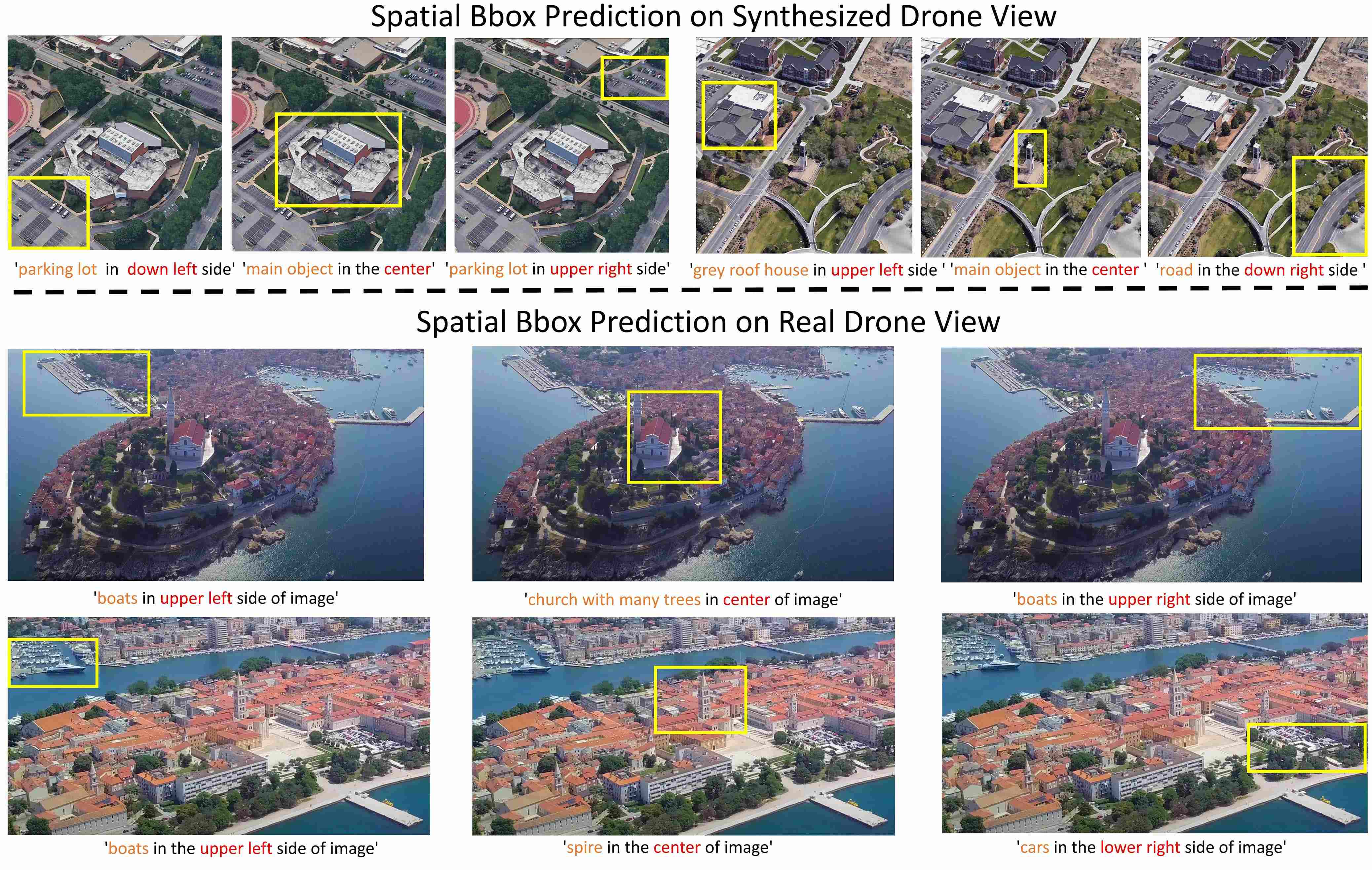}
    \vspace{-.1in}
    \caption{\textbf{Bounding box prediction on unseen images.} We evaluate images from both synthesized and real drone views in the wild. Our approach predicts correct regions even though there exist many similar instances in the entire scene. 
    It shows the necessity of the proposed spatial relation matching.
}
\label{fig:model_retrieval}\vspace{-.1in}
\end{figure*}

\begin{figure}[t]
	\centering
	\begin{minipage}{0.5\linewidth}
		\includegraphics[width=.98\linewidth]{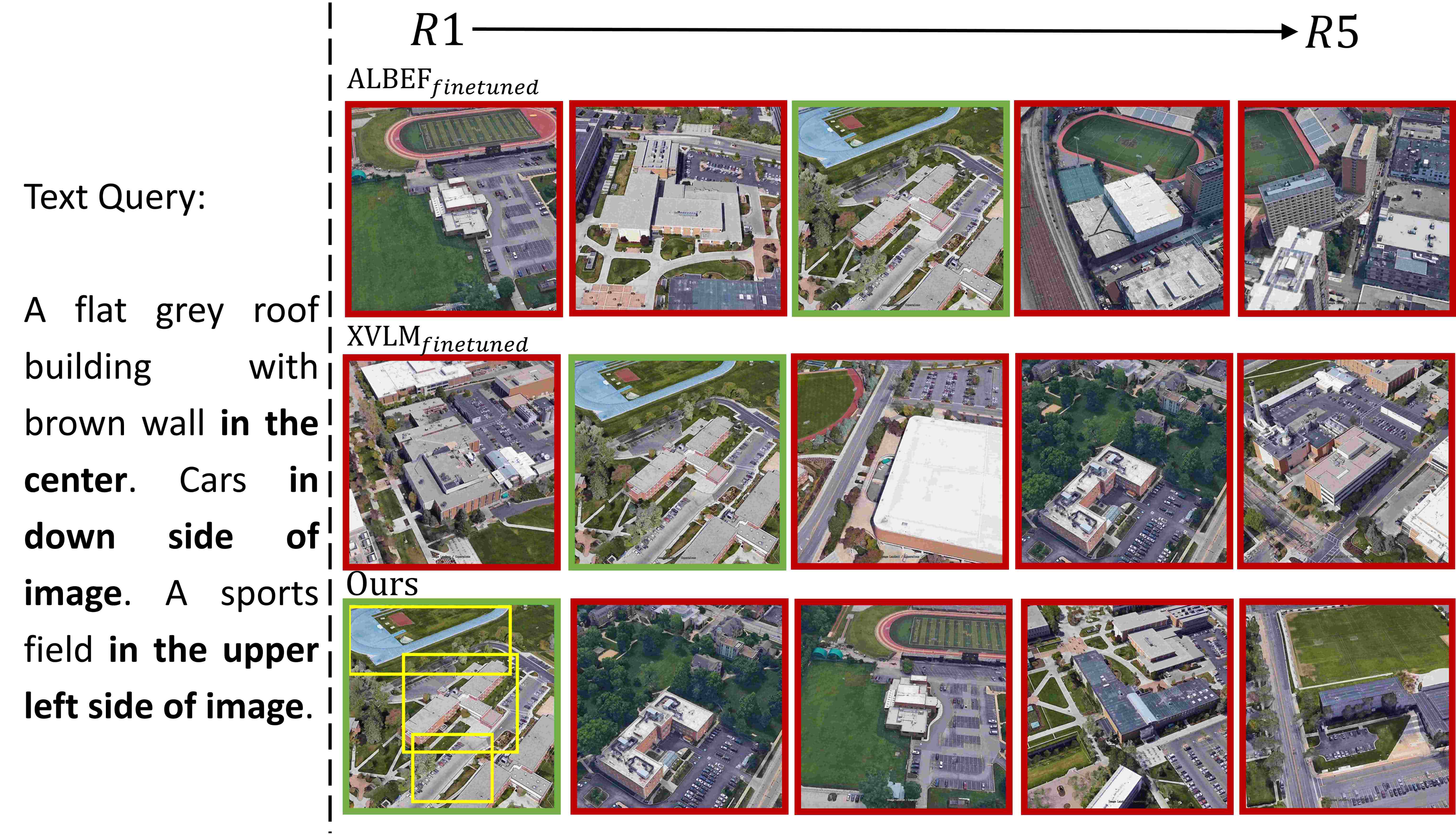}
          \vspace{0.15in}
	\end{minipage}%
	\begin{minipage}{0.49\linewidth}
		\caption{\textbf{Qualitative text-to-image retrieval results.} Here we compare our method with two baselines. The ranking list is in descending order from left to right according to the similarity score.
The images in red boxes are false-matched, while the green ones are true-matched. The keywords are highlighted in \textbf{bold}.
        }
		\label{fig:text query}
	\end{minipage}
 \vspace{-.1in}
\end{figure}

\noindent\textbf{Spatial Text Grounding.}\label{sec:grounding} We further evaluate our spatial bounding box prediction 
on both synthesized and drone-view images in the wild (see Fig.~\ref{fig:model_retrieval}).
(1) It shows the strength in spatial matching, not just on familiar, trained images but also on new, real-world scenes. 
For instance, buildings and objects on the sea are never included in our training data, but the model could easily capture the boats and buildings based on our text instruction which indicates that the model has the potential to handle real-world navigation tasks.
(2) The images also accentuate the robustness in discerning between objects solely based on textual descriptions that define their spatial relationships, even when multiple instances of the same object are present within the same image.
For example, when two parking lots are shown in the synthesized image, the model could detect the proposed parking lot based on the spatial word we provided. Also, as shown in the real drone image, when a harbour with boats on both sides, the model could also capture the proposed object based on the instruction. This level of fine-grained discrimination emphasizes the understanding of spatial language, accurately mapping words that convey spatial relationships to the specific regions of the image they describe.

\noindent\textbf{Text Query Retrieval.}
As shown in Fig.~\ref{fig:text query}, our method shows spatial-aware capabilities, achieving a higher recall compared to baseline models.
Spatial descriptors enable accurate image identification based not only on object labels but also on the integration of spatial relations. 
For instance, the keywords, \eg, ``in the center'', ``in the down side'', and ``in upper left'', are well captured by our learned model. These keywords help our model to find the object of interest. %, which enhances the recall rate in the whole retrieval process.
The results in the top rows show that the baseline still could retrieve the content-similar image, \eg, a car, sports field or colour, but they miss the spatial alignment, which is common in real-world scenarios.

\section{Conclusion}
In this work, we introduce GeoText-1652, a new vision-language dataset that enhances natural language-guided drone geolocalization, addressing the challenges of dataset availability and alignment of language with fine-grained visual representations. The dataset enables two tasks: text-to-image and image-to-text retrieval for precise drone navigation and target localization. 
We also introduce a new blending spatial matching, leveraging region-level relationships between drone-view images and textual descriptions. The proposed method outperforms other cross-modality approaches in recall accuracy and shows good generalization in real-world scenarios.

\section*{Acknowledgement}
The paper is supported by Start-up Research Grant at the University of Macau (SRG2024-00002-FST).

\bibliographystyle{splncs04}
\bibliography{egbib}

\end{document}